\documentclass[a4paper,english,oneside,reqno]{amsart}
\usepackage[T1]{fontenc}
\usepackage[latin9]{inputenc}
\usepackage{amsthm}
\usepackage{amstext}
\usepackage{setspace}
\usepackage{amssymb}
\makeatletter
\numberwithin{equation}{section} %% Comment out for sequentially-numbered
\numberwithin{figure}{section} %% Comment out for sequentially-numbered
\theoremstyle{plain}
\theoremstyle{plain}
\newtheorem{thm}{Theorem}
  \theoremstyle{plain}
  \newtheorem{lem}[thm]{Lemma}

\usepackage[colorlinks=true, pdfstartview=FitV, linkcolor=blue,citecolor=blue, urlcolor=blue]{hyperref}
\usepackage[numbers,sort]{natbib}

\makeatother

\usepackage{babel}

\begin{document}

\title{How Random are a Learner's Mistakes ?}

\author{Joel Ratsaby}

\address{Department of Electrical and Electronics Engineering, Ariel University Center of Samaria, Ariel 40700, ISRAEL. \emph{Email}:
ratsaby@ariel.ac.il}
\begin{abstract}
Given a random binary sequence $X^{(n)}$ of random variables, $X_{t},$
$t=1,2,\ldots,n$, for instance, one that is generated by a Markov
source of order $k^{*}$ (each state represented by $k^{*}$ bits).
Let $\beta$ be the probability of $"X_{t}=1"$ and assume it is constant
with respect to $t$ (due to stationarity). Consider a learner based
on a parametric model, for instance a Markov model of order $k$,
who trains on a sample sequence $x^{(m)}$ which is randomly drawn
by the source. Test the learner's performance by giving it a sequence
$x^{(n)}$ (generated by the source) and check its predictions on
every bit of $x^{(n)}.$ An error occurs at time $t$ if the  prediction
$Y_{t}$ differs from the true bit value $X_{t}$. Denote by $\xi^{(n)}$
the sequence of errors where the error bit $\xi_{t}$ at time $t$
equals $1$ or $0$ according to whether the event of an error occurs
or not, respectively. Consider the subsequence $\xi^{(\nu)}$ of $\xi^{(n)}$
which corresponds to the errors of predicting a $0$, i.e., $\xi^{(\nu)}$
consists of the bits of $\xi^{(n)}$ only at times $t$ such that
$Y_{t}=0.$ In this paper we compute an upper bound on the deviation
of the frequency of $1$s of $\xi^{(\nu)}$ from $\beta$ showing
dependence on $k$, $m$, $\nu$. 
\end{abstract}



\maketitle

\section{\label{sec1}Overview}

Let $\left\{ X_{t}:t=1,2,\ldots\right\} $ be a sequence of Bernoulli
random variables possessing the following Markovian property,

\begin{equation}
P\left(X_{t}=x\mid X_{1}=x_{1},\ldots,X_{t-1}=x_{t-1}\right)=P\left(X_{t}=x\mid X_{t-k}=x_{t-k},\ldots,X_{t-1}=x_{t-1}\right)\label{eq:mm}\end{equation}
for some fixed $k$ and where $x_{t-k},\ldots,x_{t-1}$, and $x$
are $0$ or $1$, and $t=k+1$, $k+2$, $\cdots$. The model is known
as the $k^{th}$ order Markov chain and we denote it by $\mathcal{M}_{k}$.
Let us define by $\mathbb{S}_{k}=\left\{ 0,1\right\} ^{k}$ the state
space of $\mathcal{M}_{k}$ and let $s^{(i)}\in\mathbb{S}_{k}$ denote
the $i^{th}$ state, $i=0,1,\ldots,2^{k}-1$. Denote by $\left\{ S_{t}:t=k,k+1,\ldots\right\} $
the sequence of random state variables where \begin{eqnarray*}
S_{t} & := & \left(X_{t-k+1},X_{t-k+2},\ldots,X_{t}\right)\end{eqnarray*}
represents the state at time $t$. Let $T_{k}=\left[t_{i,j}\right]$,
$1\leq i,j\leq2^{k},$ be a state-transition matrix associated with
model $\mathcal{M}_{k}$ with elements $t_{i,j}:=P(S_{t+1}=s^{(j)}\mid S_{t}=s^{(i)})$,
$0\leq i,j\leq2^{k}-1$. Clearly, the structure of $\mathcal{M}_{k}$
allows for only two possible outgoing transitions from a state $S_{t}$
to the next state $S_{t+1}$ since $S_{t+1}$ can take only one of
the two values $\left(X_{t-k+2},\ldots,X_{t},0\right)$ or $\left(X_{t-k+2},\ldots,X_{t},1\right)$.
We call them \emph{type-0} and \emph{type-1} transitions. Using $\mathcal{M}_{k}$
to produce a random sequence is done by randomly drawing a state sequence
and outputting $1$ or $0$ for each type-$1$ or type-$0$ transition,
respectively.

In this paper we consider the following estimation problem: 

\emph{Estimation problem}: A source produces a \emph{data} string
$x^{(m+n)}=\left(x_{1},\ldots,x_{m+n}\right)$, by randomly drawing
$m+n$ consecutive bits according to a Markov chain $\mathcal{M}_{k^{*}}$
of order $k^{*}$. A learner (not knowing the value of $k^{*}$) estimates
the parameters of another model $\mathcal{M}_{k}$ based on the initial
subsequence $x^{(m)}=\left(x_{1},\ldots,x_{m}\right)$ which is called
\emph{training} sequence. Afterwards, the learner uses $\mathcal{M}_{k}$
to make a prediction $y_{t}$ for each of the remaining bits $x_{m+t}$,
$t=1,2,\ldots,n$, which form the \emph{testing} sequence $x^{(n)}=\left(x_{m+1},\ldots,x_{m+n}\right)$.
Denote by $\xi^{(n)}=\left\{ \xi_{t}\right\} _{t=1}^{n}$ the corresponding
binary sequence of mistakes where $\xi_{t}=1$ if $y_{t}\neq x_{m+t}$
and is $0$ otherwise. Denote by $\xi^{(\nu)}=\left\{ \xi_{i_{j}}\right\} _{j=1}^{\nu}$,
$\nu\geq\ell$, a subsequence of $\xi^{(n)}$ with time instants $i_{j}$
corresponding to $0$-predictions, $y_{i_{j}}=0$, $1\leq j\leq\nu$.
Note that $\xi^{(\nu)}$ is also a subsequence of the input sequence
$x^{(n)}$ hence effectively the learner acts as a selection rule
$\Gamma_{d}$ which picks certain bits $\xi^{(\nu)}$ from $x^{(n)}$
according to an algorithm that specifies its prediction rule. In this
paper we compute a large deviation bound for the error sequence $\xi^{(\nu)}$.

\section{Introduction}

From basic theory on finite Markov chains, since the matrix $T:=T_{k^{*}}$
 is stochastic (i.e., the sum of the elements in any row equals $1$)
then $\mathcal{M}_{k^{*}}$ has a stationary joint probability distribution
\begin{eqnarray*}
\mathbb{P}\left(X^{(n)}\right) & := & P(X_{1},\ldots,X_{n})\end{eqnarray*}
which is not necessarily unique. To keep the notation simple we use
$\mathbb{P}$ to denote also any marginal distribution derived from
the stationary joint distribution. For instance, $\mathbb{P}\left(X_{1},X_{2},X_{3}\right)=\sum_{\left(x_{4},\ldots,x_{n}\right)}\mathbb{P}\left(X_{1},X_{2},X_{3},X_{4}=x_{4}\ldots,X_{n}=x_{n}\right)$.
Henceforth, all random binary sequences are assumed to be drawn according
to this probability distribution $\mathbb{P}$. Thus for any $k$
and $\ell$ satisfying $\ell\geq k\geq1$ the probability of a string
$x^{(\ell)}=\left(x_{1},x_{2},\ldots,x_{\ell}\right)$ can be expressed
as \begin{eqnarray}
\lefteqn{\mathbb{P}(X_{1}=x_{1},X_{2}=x_{2},\ldots X_{\ell}=x_{\ell})}\nonumber \\
 & = & \mathbb{P}\left(S_{k}=\left(x_{1},\ldots x_{k}\right)\right)\mathbb{P}\left(S_{k+1}=\left(x_{2},\ldots x_{k+1}\right)\biggl|S_{k}=\left(x_{1},\ldots x_{k}\right)\right)\nonumber \\
 &  & \cdots\mathbb{P}\left(S_{\ell}=\left(x_{\ell-k+1},\ldots x_{\ell}\right)\biggl|S_{\ell-1}=\left(x_{\ell-k},\ldots x_{\ell-1}\right)\right).\label{eq:P}\end{eqnarray}
 Let us denote by \begin{equation}
\beta=\mathbb{P}\left(X_{t}=1\right)\label{eq:beta}\end{equation}
the stationary probability of the event $\left\{ X_{t}=1\right\} $
at time $t=1,2,\ldots$ .

\emph{Data generation}: We henceforth assume that the source reached
stationarity and produces the data sequence $x^{(m+n)}$ with respect
to $\mathbb{P}$. 

Consider the learner's model $\mathcal{M}_{k}$. Its set of parameters
are the true (unknown) probability values of transitions between states
in $\mathbb{S}_{k}$ where the probability values are assigned according
to the source distribution $\mathbb{P}$. We denote them by \[
p_{ij}:=\mathbb{P}(S_{t+1}=s^{(j)}\mid S_{t}=s^{(i)}),\quad s^{(i)},s^{(j)}\in\mathbb{S}_{k}.\]
For instance, suppose $k^{*}=3$ and $k=2$ and consider two states
$s^{(i)}=\left(0,1\right)$ and $s^{(j)}=\left(1,1\right)$. The corresponding
transition probability is \[
p_{i,j}=P\left(\left(1,1\right)\mid\left(0,1\right)\right)=\frac{\mathbb{P}\left(X_{t-1}=0,X_{t}=1,X_{t+1}=1\right)}{\mathbb{P}\left(X_{t-1}=0,X_{t}=1\right)}.\]
Based on $x^{(m)}$ the learner estimates $p_{i,j}$ by \[
\hat{p}_{ij}=\frac{m_{i,j}}{m_{i}}\]
where for a state $s^{(i)}\in\mathbb{S}_{k}$, $m_{i}$ denotes the
number of times that $s^{(i)}$ appears in $x^{(m)}$ and $m_{i,j}$
denotes the number of times there is a transition from state $s^{(i)}$
to $s^{(j)}$ in $x^{(m)}$. For instance, if $k=3$, $x^{(m)}=011010101$
and $s^{(i)}=101$ then $m_{i}=3$. Thus $\hat{p}_{ij}$ are the frequency
of state-transitions in $x^{(m)}$. Note that $m_{i}$, $0\leq i\leq2^{k}-1$,
are dependent random variables since the Markov chain may visit each
state a random number of times and they must satisfy $\sum_{i=0}^{2^{k}-1}m_{i}=m-k+1$. 

After training, the learner is tested on the remaining $n$ bits of
the data $x^{(n)}=x_{m+1},x_{m+2},\ldots,x_{m+n}$ . It makes a binary
prediction $Y_{t}$ for $X_{t}$, $t=m+1,\ldots,n$ based on the maximum
\emph{a posteriori} probability which is defined as follows: suppose
that the current state is $s^{(i)}\in\mathbb{S}_{k}$ then the prediction
is

\begin{equation}
\hat{d}(i):=\left\{ \begin{array}{cc}
1 & \text{if }\hat{p}(1|i)>1-\hat{p}(1|i)\\
0 & \text{otherwise},\end{array}\right.\label{eq:ruled}\end{equation}
where $\hat{p}(1|i)$ is defined as $\hat{p}_{ij}$ for the state
$s^{(j)}$ obtained from $s^{(i)}$ by a type-1 transition, i.e.,
if $s^{(i)}=\left(x_{\ell+1},x_{\ell+2},\ldots,x_{\ell+k}\right)$
then $s^{(j)}=\left(x_{\ell+2},x_{\ell+3},\ldots,x_{\ell+k},1\right)$.
The corresponding true probability value is denoted by $p(1|i)=p_{ij}$.
Note that (\ref{eq:ruled}) may be expressed alternatively as\begin{equation}
\hat{d}(i)=\left\{ \begin{array}{cc}
1 & \text{if }\hat{p}(1|i)>\frac{1}{2}\\
0 & \text{otherwise}.\end{array}\right.\label{eq:rule12}\end{equation}

We claim that $\hat{p}(1|i)$, $0\leq i\leq2^{k}-1$, are independent
random variables when conditioned on the vector $\underline{m}:=[m_{0},\ldots,m_{2^{k}-1}]$.
We now prove the claim which will be used in Section \ref{sec:Proof-of-Theorem}.
Let us denote by $\sigma^{(m)}=\left(\sigma_{k},\ldots,\sigma_{m}\right)$,
$\sigma_{i}\in\mathbb{S}_{k}$, $k\leq i\leq m$, the particular sequence
of states corresponding to the sequence $x^{(m)}$. To show the dependence
of $x^{(m)}$ on $\sigma^{(m)}$ we will sometimes write $x^{(m)}=x\left(\sigma^{(m)}\right)$.
Then by (\ref{eq:P}) we have\begin{eqnarray*}
\lefteqn{\mathbb{P}\left(X_{1}=x_{1},\ldots,X_{m}=x_{m}\right)}\\
 & = & \mathbb{P}\left(S_{k}=\sigma_{k}\right)\mathbb{P}\left(S_{k+1}=\sigma_{k+1}\biggr|S_{k}=\sigma_{k}\right)\mathbb{P}\left(S_{k+2}=\sigma_{k+2}\biggr|S_{k+1}=\sigma_{k+1}\right)\\
 &  & \cdots\mathbb{P}\left(S_{m}=\sigma_{m}\biggr|S_{m-1}=\sigma_{m-1}\right).\end{eqnarray*}
Since at every bit there are only two types of transitions then not
every sequence $\sigma^{(m)}\in(\mathbb{S}_{k})^{m-k+1}$ is possible.
For instance, if $k=3$ then the state sequence $\left(010,100,001\right)$
is valid but $\left(010,110,001\right)$ is not valid. Denote by $V\subset(\mathbb{S}_{k})^{m-k+1}$
the set of \emph{valid} state sequences $\sigma^{(m)}$. We now show
that if $\sigma^{(m)}$ is in $V$ then, conditioned on $\underline{m}$,
any other state sequence that visits the same states as $\sigma^{(m)}$
the same number of times (perhaps in a different order) must have
the same probability. For any state $s^{(i)}\in\mathbb{S}_{k}$ denote
by $N(1|i)$ the random variable whose value is the number of type-1
transitions from state $s^{(i)}$ in a sequence of random states $S^{(m)}=S_{k},S_{k+1},\ldots,S_{m}$.
Define by $N_{\sigma^{(m)}}(1|i)$ the number of type-1 transitions
from state $s^{(i)}$ in the sequence $\sigma^{(m)}$. Since all state
transitions are either type-0 or type-1  then we have\begin{eqnarray}
\lefteqn{\mathbb{P}\left(X^{(m)}=x\left(\sigma^{(m)}\right)\biggl|\underline{m},S^{(m)}=\sigma^{(m)},\sigma^{(m)}\in V\right)}\nonumber \\
 & = & \prod_{0\leq i\leq2^{k}-1}\left(p(1|i)\right)^{N_{\sigma^{(m)}}(1|i)}\left(1-p(1|i)\right)^{m_{i}-N_{\sigma^{(m)}}(1|i)}\label{eq:prod1}\end{eqnarray}
where $p(1|i)$ was defined above. Let $\alpha$ be a non-negative
integer parameter and define the random variable $N(i):=\left[N(1|i),\alpha-N(1|i)\right]$.
Associate a conditional probability function with parameter $\alpha$
for the random variable $N(i)$ as \[
\mathbb{P}\left(N(i)=\left[\ell,\alpha-\ell\right]\biggl|\alpha\right)=\left(p(1|i)\right)^{\ell}\left(1-p(1|i)\right)^{\alpha-\ell}.\]
Then the right side of (\ref{eq:prod1}) equals\begin{equation}
\prod_{0\leq i\leq2^{k}-1}\mathbb{P}\left(N(i)=\left[N_{\sigma^{(m)}}\left(1|i\right),m_{i}-N_{\sigma^{(m)}}\left(1|i\right)\right]\right).\label{eq:Ni}\end{equation}
For a fixed value of $m_{i}$ the event {}``$N(i)=\left[N_{\sigma^{(m)}}\left(1|i\right),m_{i}-N_{\sigma^{(m)}}\left(1|i\right)\right]$''
is equivalent to the event {}``$\hat{p}(1|i)=\frac{N_{\sigma^{(m)}}(1|i)}{m_{i}}$''.
Hence alternatively, the right side of (\ref{eq:Ni}) can be expressed
as \begin{equation}
\prod_{0\leq i\leq2^{k}-1}\mathbb{P}\left(\hat{p}(1|i)=\frac{N_{\sigma^{(m)}}(1|i)}{m_{i}}\right).\label{eq:p1m}\end{equation}
The right side of (\ref{eq:p1m}) is a product of probability functions
of the random variables $\hat{p}(1|i)$. So conditioned on $\underline{m}$
and on the event that $x^{(m)}$ corresponds to a valid state sequence
$\sigma^{(m)}$, the event that $x^{(m)}$ is generated by the source
Markov chain $\mathcal{M}_{k^{*}}$ is equivalent to the event that
its corresponding state sequence $\sigma^{(m)}$ has transition frequencies
$\hat{p}(1|i)$ that independently take the particular values $\frac{N_{\sigma^{(m)}}(1|i)}{m_{i}}$
as prescribed in $x^{(m)}$. The claim is proved. It also follows
that $\hat{p}(1|i)$ is the average of independent Bernoulli trials
(success taken as a type-1 transition from state $s^{(i)}$). It is
distributed according to the Binomial distribution with parameters
$m_{i}$ and $p(1|i)$.

We now summarize the problem setting under which the main result of
the paper holds.

\emph{Problem setting}: Let $0<\delta<1$ and $k,\ell,m,n$ be positive
integers. Let $\mathbb{P}$ be the stationary probability distribution
based on a finite, ergodic and reversible Markov chain with probability-transition
matrix $T$ that has a second largest eigenvalue $\lambda$. All probability
values are measured according to $\mathbb{P}$. Denote by $\gamma=(1-\max\left\{ 0,\lambda\right\} )/(1+\max\left\{ 0,\lambda\right\} )$.
After reaching stationarity the source generates a binary sequence
$X^{(n)}=X_{1},X_{2,}\ldots,X_{n}$ by repeatedly drawing $X_{t}$
according to $\mathbb{P}$ . Denote by $\beta=\mathbb{P}\left(X_{t}=1\right)$.
Let $x^{(m+n)}$ be a data-sequence obtained by randomly drawing according
to $\mathbb{P}$. Let the learner's model $\mathcal{M}_{k}$ be Markov
of order $k$, and denote by $p(1|i)$ the probability of making a
type-1 transition from state $s^{(i)}$ of $\mathcal{M}_{k}$ . The
learner uses the first $m$ bits, $x^{(m)}$, to estimate $p(1|i$)
by $\hat{p}(1|i)$. Let $m_{i}$ denote the number of times that state
$s^{(i)}$ appears in $x^{(m)}$, $\sum_{0\leq i\leq2^{k}-1}m_{i}=m-k+1$.
After training, the learner's decision at state $s^{(i)}$ is to output
$1$ if $\hat{p}(1|i)>\frac{1}{2}$ else output $0$. Denote by $\mu_{i}$
the probability that a Binomial random variable with parameters $m_{i}$,
$p(1|i)$, is larger (or smaller) than $\frac{m_{i}}{2}$ given that
$p(1|i)$ is smaller (or larger) than $\frac{1}{2}$, respectively.
Let $\mu=\frac{1}{2^{k}}\sum_{0\leq i\leq2^{k}-1}\mu_{i}$. Let $\rho^{(m)}=\frac{1}{2^{k}}\sum_{0\leq i\leq2^{k}-1}\exp\left\{ -2m_{i}\left(\frac{1}{2}-p(1|i)\right)^{2}\right\} $.
Using $\mathcal{M}_{k}$ the learner is tested incrementally on the
remaining $n$ bits $x^{(n)}=x_{m+1},\ldots,x_{m+n}$ of the data
and predicts an output bit $y_{t}$ for bit $x_{t}$ in $x^{(n)}$
to be $1$ if $\hat{p}(1|i)>\frac{1}{2}$, else $0$. Denote by $\xi^{(n)}$
the sequence of mistakes where $\xi_{t}=1$ if $y_{t}\neq x_{t}$,
and $\xi_{t}=0$ otherwise, $m+1\leq t\leq m+n$. Denote by $\xi^{(\nu)}=\left\{ \xi_{i_{j}}\right\} _{j=1}^{\nu}$,
$\nu\geq\ell$, the subsequence of $\xi^{(n)}$ with time instants
$i_{j}$ corresponding to $0$-predictions, $y_{i_{j}}=0$, $1\leq j\leq\nu$.
Note that $\xi^{(\nu)}$ is also a subsequence of the input sequence
$x^{(n)}$ hence effectively the learner acts as a selection rule
which picks certain bits $\xi^{(\nu)}$ from $x^{(n)}$.

Let 

\begin{eqnarray*}
\epsilon^{2}(\ell,\gamma,\delta,k,m) & := & \frac{1}{2\ell\gamma}\Biggl[2^{k}\left(\sqrt{\frac{1}{2^{k-2}}\ln\left(\frac{2}{\delta}\right)}+\rho^{(m)}\right)\ln\left(\frac{e}{\left(\sqrt{\frac{1}{2^{k-2}}\ln\left(\frac{2}{\delta}\right)}+\rho^{(m)}\right)}\right)\\
 & + & \ln2+\ln\left(\frac{4}{\delta}\right)\Biggr]\end{eqnarray*}

and assume that the learner's model order $k$ satisfies, 

\[
k\geq2+2\log_{2}\left(\frac{1}{(2e-1)\mu}\right)+\log_{2}\left(\ln\left(\frac{2}{\delta}\right)\right).\]
We now state the main result of the paper.
\begin{thm}
\label{Th1}For any $0<\delta<1$, with probability at least $1-\delta$
the deviation between $\beta$ and the frequency of $1$s of the sequence
$\xi^{(\nu)}$ is bounded as \[
\left|\frac{1}{\nu}\sum_{j=1}^{\nu}\xi_{j}^{(\nu)}-\beta\right|\leq\epsilon(\ell,\gamma,\delta,k,m).\]

\end{thm}
Before presenting the proof we make the following remarks,
\begin{enumerate}
\item The effect of the training sequence length $m$ on $\epsilon$ is
as $O(\rho^{(m)})$ which is $O(e^{-m})$. As $m$ increases the class
of possible learnt models (hypothesis class) decreases in size thereby
decreasing the bound $\epsilon$ on the deviation of the error sequence.
\item The effect of the learner's model order $k$ is opposite of that of
$m$. We see that $\epsilon=O(2^{k/2})$ and as $k$ increases, the
hypothesis class increases in size.
\item The effect of the length $\ell$ of the error sequence on $\epsilon$
is as $O(\frac{1}{\ell})$. Clearly, the longer the subsequence the
less chance that its frequency of 1s deviate from the mean $\beta$.
\item The effect of the inter-dependence between the states of the source
model $\mathcal{M}_{k^{*}}$ on $\epsilon$ is as $O(\frac{1}{\gamma})$.
As the dependence increases, $\gamma$ decreases which increases the
possible deviation size $\epsilon$. As $\gamma$ decreases, the bits
of the sequence $X^{(n)}$ become less dependent and $\epsilon$ decreases.
\end{enumerate}

\section{\label{sec:Proof-of-Theorem}Proof of Theorem \ref{Th1}}

A prediction decision-rule is denoted by a binary vector \begin{equation}
d:=[d(0),\ldots,d(2^{k}-1)].\label{eq:d}\end{equation}
Note that $d$ describes the prediction made by the learner at each
state of the model.

Assume that the length $m$ of the training sequence $x^{(m)}$ is
fixed. Let us define by $d^{*}$ the Bayes optimal decision. Clearly,
$d^{*}(i)=1$ when $p(1|i)>\frac{1}{2}$ and $d^{*}(i)=0$ otherwise,
$1\leq i\leq2^{k}$. Let us define the following set for $0\leq r\leq2^{k}$,
\begin{equation}
A_{r}^{(k)}=\left\{ d\in\left\{ 0,1\right\} ^{2^{k}}:\left\Vert d-d^{*}\right\Vert \leq r\right\} ,\label{eq:A}\end{equation}
where $\left\Vert d\right\Vert $ denotes the $l_{1}$-norm of $d$.
Consider the Bernoulli random variable $\chi_{i}$ which equals $1$
if $\hat{d}(i)\neq d^{*}(i)$ and $0$ otherwise ($\hat{d}$ is defined
in (\ref{eq:ruled})). The event that $\chi_{i}=1$ occurs if $\hat{p}(1|i)>\frac{1}{2}$
and $p(1|i)\leq\frac{1}{2}$ or if $\hat{p}(1|i)\leq\frac{1}{2}$
and $p(1|i)>\frac{1}{2}$. Since $\hat{p}(1|i)$ is the average of
$m_{i}$ i.i.d. Bernoulli random variables each with an expected value
of $p(1|i)$ then we have in the case of $p(1|i)\leq\frac{1}{2}$
that\begin{eqnarray*}
\mathbb{P}\left(\hat{p}(1|i)>\frac{1}{2}\right) & = & \mathbb{P}\left(\hat{p}(1|i)-p(1|i)>\frac{1}{2}-p(1|i)\right)\\
 & = & \mathbb{P}\left(\hat{p}(1|i)>p(1|i)+\alpha_{i}\right)\end{eqnarray*}
where \begin{eqnarray*}
\alpha_{i} & = & \frac{1}{2}-p(1|i).\end{eqnarray*}
By Chernoff's bound \cite{Chernoff52} we have \[
\mathbb{P}\left(\hat{p}(1|i)>p(1|i)+\alpha_{i}\right)\leq\exp\left\{ -2m_{i}\alpha_{i}^{2}\right\} .\]
Similarly, if $p(1|i)>\frac{1}{2}$ then, denoting by $\alpha_{i}=p(1|i)-\frac{1}{2}$,
we have \begin{eqnarray*}
\mathbb{P}\left(\hat{p}(1|i)\leq\frac{1}{2}\right) & = & \mathbb{P}\left(\hat{p}(1|i)-p(1|i)\leq\frac{1}{2}-p(1|i)\right)\\
 & = & \mathbb{P}\left(\hat{p}(1|i)\leq p(1|i)-\alpha_{i}\right)\\
 & \leq & \exp\left\{ -2m_{i}\alpha_{i}^{2}\right\} .\end{eqnarray*}
Therefore, regardless of the value of $p(1|i)$, we have \begin{equation}
\mu_{i}:=\mathbb{P}\left(\chi_{i}=1\right)\leq\rho_{i}^{(m)}\label{eq:bnd1}\end{equation}
where \begin{eqnarray*}
\rho_{i}^{(m)}: & = & \exp\left\{ -2m_{i}\alpha_{i}^{2}\right\} ,\;1\leq i\leq2^{k}.\end{eqnarray*}
As shown in the previous section, conditioned on $\underline{m}$,
the $\hat{p}(1|i)$ are independent. Hence $\left\{ \chi_{i}\right\} _{i=1}^{2^{k}}$
are independent non-identically distributed Bernoulli random variables
(known as Poisson trials).

According to $\mathbb{P}$ the probability of the event that $\hat{d}$
is not in $A_{r}^{(k)}$ is the same as the probability of this event
conditioned on the state sequence $\sigma^{(m)}$ being valid. Hence,
\begin{eqnarray}
\mathbb{P}\left(\hat{d}\not\in A_{r}^{(k)}\right) & = & \mathbb{P}\left(\hat{d}\not\in A_{r}^{(k)}\biggl|\:\sigma^{(m)}\in V\right)\nonumber \\
 & = & \sum_{\underline{m}}\mathbb{P}\left(\hat{d}\not\in A_{r}^{(k)}\biggl|\underline{m},\:\sigma^{(m)}\in V\right)\mathbb{P}\left(\underline{m}\biggl|\sigma^{(m)}\in V\right)\label{eq:bound1}\end{eqnarray}
where the sum runs over all non-negative $\underline{m}$ that satisfy
$\sum_{0\leq i\leq2^{k}-1}m_{i}=m-k+1$. We now bound the first factor
inside the sum by a quantity which only depends on $m$ (not on the
specific vector $\underline{m}$). Denote by $\hat{\chi}=\frac{1}{2^{k}}\sum_{i=1}^{2^{k}}\chi_{i}$,
and recall the definitions $\mu:=\frac{1}{2^{k}}\sum_{i=1}^{2^{k}}\mu_{i}$
and $\rho^{(m)}:=\frac{1}{2^{k}}\sum_{i=1}^{2^{k}}\rho_{i}^{(m)}$.
From (\ref{eq:bnd1}) it follows that $\mu\leq\rho^{(m)}$. Conditioned
on $\underline{m},\:\sigma^{(m)}\in V$ we have for any $\epsilon>0$,

\begin{eqnarray*}
\mathbb{P}\left(\hat{d}\not\in A_{r}^{(k)}\right) & = & \mathbb{P}\left(\hat{\chi}>\frac{r}{2^{k}}\right)\\
 & = & \mathbb{P}\left(\hat{\chi}-\mu>\frac{r}{2^{k}}-\mu\right)\\
 & \leq & \mathbb{P}\left(\hat{\chi}-\mu>\frac{r}{2^{k}}-\rho^{(m)}\right).\end{eqnarray*}
We will use the following lemma which bounds the deviation of the
average of Poisson trials from their mean.
\begin{lem}
\label{lem:Let--be}Let $X_{1},\ldots,X_{n}$ be independent Bernoulli
random variables with $P(X_{i}=1)=\mu_{i}$ and denote by $\mu=\frac{1}{n}\sum_{i=1}^{n}\mu_{i}.$
Then for any $0<\gamma\leq(2e-1)\mu$ the following bound holds:
\end{lem}
\[
P\left(\frac{1}{n}\sum_{i=1}^{n}X_{i}>\mu+\gamma\right)\leq e^{-n\gamma^{2}/4}.\]
The proof of the lemma is based on applying Chernoff bound on the
tail probability of the sum of Poisson trials (similar to Theorem
4.1 and 4.3, in \cite{Motwani95}). 

Substituting for $\gamma$ the value $\frac{r}{2^{k}}-\rho^{(m)}$
in the above lemma and recalling the theorem's condition that \[
k\geq2+2\log_{2}\left(\frac{1}{(2e-1)\mu}\right)+\log_{2}\left(\ln\left(\frac{2}{\delta}\right)\right)\]
which, with the following choice for $r$, 

\begin{equation}
r=2^{k}\left(\sqrt{\frac{1}{2^{k-2}}\ln\left(\frac{2}{\delta}\right)}+\rho^{(m)}\right),\label{eq:rr}\end{equation}
ensures that $\frac{r}{2^{k}}-\rho^{(m)}\leq(2e-1)\mu$, yields the
following bound, \begin{equation}
\mathbb{P}\left(\hat{d}\not\in A_{r}^{(k)}\right)\leq e^{-2^{k-2}\left(\frac{r}{2^{k}}-\rho^{(m)}\right)^{2}}.\label{eq:dnotA}\end{equation}
Next, we estimate the cardinality of the set $A_{r}^{(k)}$. Without
loss of generality let $d^{*}=\left[0,0,\ldots,0\right]$ then we
have \begin{eqnarray}
\left|A_{r}^{(k)}\right| & = & \sum_{i=0}^{r}{2^{k} \choose i}\nonumber \\
 & \leq & \left(\frac{e2^{k}}{r}\right)^{r}.\label{eq:abou}\end{eqnarray}
Since the error subsequence $\xi^{(\nu)}$ is also a subsequence of
$x^{(n)}$ then we associate a selection rule $\Gamma_{d}:\left\{ 0,1\right\} ^{n}\rightarrow\left\{ 0,1\right\} ^{\nu}$
which \emph{selects} $\xi^{(\nu)}$ from $x^{(n)}$. Let $E_{d,\epsilon}^{(\ell)}$
denote the event that based on a given $\Gamma_{d}$ the selected
subsequence $\xi^{(\nu)}$ is of length at least $\ell$ and its frequency
of $1$s deviates from the expected value $\beta$ by at least $\epsilon$.
Formally, this is defined as the large-deviation event

\[
E_{d,\epsilon}^{(\ell)}=\left\{ x^{(n)}:\,\xi^{(\nu)}=\Gamma_{d}\left(x^{(n)}\right),\,\nu\geq\ell,\,\left|\frac{\|\xi^{(\nu)}\|}{\nu}-\beta\right|>\epsilon\right\} ,\]
where $\|\xi^{(\nu)}\|$ denotes the number of $1$s in the binary
sequence $\xi^{(\nu)}$ of length $\nu$. We wish to bound from above
the probability of $E_{d,\epsilon}^{(\ell)}$.

We use the following lemma which states a rate on the strong law of
large numbers for a Markov Chain.
\begin{lem}
\label{lem32}\emph{\cite{LeonPerron04}} Let $Z_{1},\ldots Z_{n}$
be a finite ergodic and reversible Markov chain in stationary state
with a second largest eigenvalue $\lambda$ and $f$ a function taking
values in $\left[0,1\right]$ such that $\mathbb{E}f\left(Z_{i}\right)=\mu$
. Denote by $\lambda_{0}=\max\left\{ 0,\lambda\right\} $ and the
stationary probability distribution $\mathbb{P}$. Then for all $\epsilon>0$
such that $\mu+\epsilon<1$, $n\geq1$ the following bound holds:
\[
\mathbb{P}\left(\sum_{i=1}^{n}Z_{i}\geq n(\mu+\epsilon)\right)\leq e^{-2n\epsilon^{2}\left(\frac{1-\lambda_{0}}{1+\lambda_{0}}\right)}.\]

\end{lem}
The lemma appears as Theorem 1 in \cite{LeonPerron04}. We note that
the bound in this lemma coincides with Hoeffding's bound and is optimal
for $\lambda\geq0$.

We now apply the lemma to the error subsequence in order to estimate
the probability of the large deviation event $E_{d,\epsilon}^{(\ell)}$.
Denote by $X^{(n)}=\left\{ X_{m+i}\right\} _{i=1}^{n}$ and $Y^{(n)}=\left\{ Y_{m+i}\right\} _{i=1}^{n}$
the sequences of random variables produced by the source according
to $\mathcal{M}_{k^{*}}$ and the predictions made by the learner,
respectively. Let $\Xi^{(\nu)}=\left\{ \Xi_{j}\right\} _{j=1}^{\nu}$
be the sequence of random variables representing the errors made when
predicting zeros, i.e., $Y_{i_{j}}=0$, $i_{j}\in\left\{ m+1,\ldots,m+n\right\} $,
$1\leq j\leq\nu$. As mentioned above, $\Xi^{(\nu)}$ is a subsequence
of $X^{(n)}$. Denote by $S^{(n)}=\left\{ S_{m+i}\right\} _{i=1}^{n}$
the sequence of consecutive states of model $\mathcal{M}_{k^{*}}$
that correspond to $X^{(n)}$. Let $\tilde{S}{}^{(\nu)}$ denote the
subsequence of $S^{(n)}$ corresponding to the subsequence $\Xi^{(\nu)}$
of $X^{(n)}$. 

We now apply Lemma \ref{lem32} to the sequence $\Xi^{(\nu)}$. In
general $S^{(\nu)}$ may be split into parts each consisting of consecutive
states $S$ of $S^{(n)}$, i.e., \begin{eqnarray*}
\tilde{S}^{(\nu)} & = & \left(\tilde{S}_{1},\ldots,\tilde{S}_{\nu}\right)\\
 & = & \left(S_{i_{1}},S_{i_{1}+1},\ldots,S_{i_{1}+r_{1}-1},S_{i_{2}},S_{i_{2}+1},\ldots,S_{i_{2}+r_{2}-1},\cdots,S_{i_{q}},S_{i_{q}+1},\ldots,S_{i_{q}+r_{q}-1}\right),\end{eqnarray*}
where $\sum_{j=1}^{q}r_{j}=\nu$ and the parts are disjoint, i.e.,
\begin{eqnarray}
i_{j}+r_{j} & < & i_{j+1}\label{eq:les}\end{eqnarray}
for $1\leq j\leq q-1$. For any state $s\in\mathbb{S}_{k}$ let the
function $f(s)$ in Lemma \ref{lem32} be the value of the least significant
bit of the binary-representation of $s$. So for $\tilde{S}_{i}$
in $\tilde{S}{}^{(\nu)}$ we have $f(\tilde{S}{}_{i})=\Xi_{i}$ and
$\mu$ in the lemma equals $\mathbb{P}(\Xi_{i}=1)$ which equals $\beta$
by (\ref{eq:beta}). Let $\lambda$ in the lemma be the second largest
eigenvalue of the source's transition matrix $T$. 

From Chebychev's inequality (see for instance \cite{Shiryaev96}),
for any $t\geq0$, the sequence $\Xi^{(\nu)}$ satisfies, \begin{eqnarray}
\mathbb{P}\left(\sum_{j=1}^{\nu}\Xi_{j}>\nu(\beta+\epsilon)\right) & = & \mathbb{P}\left(\sum_{j=1}^{\nu}f(\tilde{S}_{j})>\nu(\beta+\epsilon)\right)\label{eq:mnm1}\\
 & \leq & \exp\left\{ -\nu t(\beta+\epsilon)\right\} \mathbb{E}\exp\left\{ t\sum_{j=1}^{\nu}f(\tilde{S}_{j})\right\} .\label{eq:exp1}\end{eqnarray}
The expectation in (\ref{eq:exp1}) is now expressed as

\begin{eqnarray}
\mathbb{E}\exp\left\{ t\sum_{j=1}^{\nu}f(\tilde{S}_{j})\right\}  & = & \mathbb{E}\exp\left\{ t\sum_{j=1}^{q-1}\sum_{k_{j}=0}^{r_{j}-1}f(S_{i_{j}+k_{j}})\right\} .\label{eq:exp5}\end{eqnarray}
Recall that $S^{(\nu)}$ is a subsequence of $S^{(n)}$ so that the
expectation is taken with respect to the joint probability distribution
of the whole sequence $S^{(n)}$. Taking into account the states not
common to $S^{(n)}$ and $\tilde{S}^{(\nu)}$ (these are at time instances
when the selection rule does not select a bit from $X^{(n)}$) and
recalling that $S^{(n)}$ starts at state $S_{m+1}$ then we write,
\begin{eqnarray*}
S^{(n)} & = & \biggl(S_{m+1},\ldots,S_{i_{1}-1},S_{i_{1}},\ldots,S_{i_{1}+r_{1}-1},S_{i_{1}+r_{1}},\ldots,S_{i_{2}-1},S_{i_{2}},\\
 &  & \ldots,S_{i_{2}+r_{2}-1},S_{i_{2}+r_{2}},\ldots,S_{i_{q}},\ldots S_{i_{q}+r_{q}-1},\ldots,S_{m+n}\biggr)\end{eqnarray*}
We can now express (\ref{eq:exp5}) explicitly as follows: \begin{eqnarray}
\lefteqn{\mathbb{E}\exp\left\{ t\sum_{j=1}^{q-1}\sum_{k_{j}=0}^{r_{j}-1}f(S_{i_{j}+k_{j}})\right\} =\sum_{s_{m+1}\in\mathbb{S}}\mathbb{P}\left(s_{m+1}\right)\sum_{s_{m+2}\in\mathbb{S}}\mathbb{P}\left(s_{m+2}\mid s_{m+1}\right)}\nonumber \\
 &  & \cdots\sum_{s_{i_{1}-1}\in\mathbb{S}}\mathbb{P}\left(s_{i_{1}-1}\mid s_{i_{1}-2}\right)\label{eq:fct0}\\
 & \cdot & \sum_{s_{i_{1}}\in\mathbb{S}}\mathbb{P}\left(s_{i_{1}}\mid s_{i_{1}-1}\right)e^{tf(s_{i_{1}})}\sum_{s_{i_{1}+1}\in\mathbb{S}}\mathbb{P}\left(s_{i_{1}+1}\mid s_{i_{1}}\right)e^{tf(s_{i_{1}+1})}\nonumber \\
 &  & \qquad\cdots\sum_{s_{i_{1}+r_{1}-1}\in\mathbb{S}}\mathbb{P}\left(s_{i_{1}+r_{1}-1}\mid s_{i_{1}+r_{1}-2}\right)e^{tf(s_{i_{1}+r_{1}-1})}\label{eq:fct1}\\
 & \cdot & \sum_{s_{i_{1}+r_{1}}\in\mathbb{S}}\mathbb{P}\left(s_{i_{1}+r_{1}}\mid s_{i_{1}+r_{1}-1}\right)\cdots\sum_{s_{i_{2}-1}\in\mathbb{S}}\mathbb{P}\left(s_{i_{2}-1}\mid s_{i_{2}-2}\right)\nonumber \\
 & \cdot & \sum_{s_{i_{2}}\in\mathbb{S}}\mathbb{P}\left(s_{i_{2}}\mid s_{i_{2}-1}\right)e^{tf(s_{i_{2}})}\sum_{s_{i_{2}+1}\in\mathbb{S}}\mathbb{P}\left(s_{i_{2}+1}\mid s_{i_{2}}\right)e^{tf(s_{i_{2}+1})}\nonumber \\
 &  & \qquad\cdots\sum_{s_{i_{2}+r_{2}-1}\in\mathbb{S}}\mathbb{P}\left(s_{i_{2}+r_{2}-1}\mid s_{i_{2}+r_{2}-2}\right)e^{tf(s_{i_{2}+r_{2}-1})}\label{eq:fct3}\\
 & \cdots\nonumber \\
 & \cdot & \sum_{s_{i_{q}}\in\mathbb{S}}\mathbb{P}\left(s_{i_{q}}\mid s_{i_{q}-1}\right)e^{tf(s_{i_{q}})}\sum_{s_{i_{q}+1}\in\mathbb{S}}\mathbb{P}\left(s_{i_{q}+1}\mid s_{i_{q}}\right)e^{tf(s_{i_{q}+1})}\nonumber \\
 &  & \qquad\cdots\sum_{s_{i_{q}+r_{q}-1}\in\mathbb{S}}\mathbb{P}\left(s_{i_{q}+r_{q}-1}\mid s_{i_{q}+r_{q}-2}\right)e^{tf(s_{i_{q}+r_{q}-1})}\nonumber \\
 & \cdot & \sum_{s_{i_{q}+r_{q}}\in\mathbb{S}}\mathbb{P}\left(s_{i_{q}+r_{q}}\mid s_{i_{q}+r_{q}-1}\right)\cdots\sum_{s_{m+n}\in\mathbb{S}}\mathbb{P}\left(s_{m+n}\mid s_{m+n-1}\right).\nonumber \end{eqnarray}

In the proof of Lemma \ref{lem32} \cite{LeonPerron04} the two-state
case ($|\mathbb{S}|=2$) is solved first. They show that factors of
the kind of (\ref{eq:fct1}) can be expressed as a product of matrices
$(MD_{t}^{2})^{r_{1}}$ where $M$ denotes the $2\times2$ transition
matrix and $D_{t}$ is a diagonal matrix $\left(\begin{array}{cc}
1 & 0\\
0 & \exp\left\{ t/2\right\} \end{array}\right)$. It follows immediately that a factor such as (\ref{eq:fct3}) equals
$(MD_{0}^{2})^{a}=M^{a}$, $a=i_{2}-i_{1}-r_{1}$. Thus, for the two
state case, the expectation (left hand side of (\ref{eq:fct0})) equals
a bilinear form $\begin{array}{cc}
[\beta & 1-\beta\end{array}]R\begin{array}{cc}
[1 & 1\end{array}]'$ where $\beta$ is the stationary probability of the first of the
two states, $v'$ denotes transpose of the vector $v$ and the square
matrix $R$ equals $M^{a_{1}}(MD_{t}^{2})^{a_{2}}M^{a_{3}}(MD_{t}^{2})^{a_{4}}\cdots M^{a_{N}}(MD_{t}^{2})^{a_{N}}$.
Since $M$ is a stochastic matrix then each of its elements is non-negative
and bounded by $1$ hence we have \begin{eqnarray}
\begin{array}{cc}
[\beta & 1-\beta\end{array}]R\begin{array}{cc}
[1 & 1\end{array}]' & \leq & \begin{array}{cc}
[\beta & 1-\beta\end{array}](MD_{t}^{2})^{a_{2}}(MD_{t}^{2})^{a_{4}}\cdots(MD_{t}^{2})^{a_{N}}\begin{array}{cc}
[1 & 1\end{array}]'\nonumber \\
 & = & \begin{array}{cc}
[\beta & 1-\beta\end{array}](MD_{t}^{2})^{\nu}\begin{array}{cc}
[1 & 1\end{array}]'.\label{eq:b11}\end{eqnarray}
Based on the proof of \cite{LeonPerron04}, multiplying (\ref{eq:b11})
by the exponential factor $\exp\left\{ -\nu t(\beta+\epsilon)\right\} $
gives an expression which is bounded from above by $\exp\left\{ -2\nu\left(\frac{1-\lambda_{0}}{1+\lambda_{0}}\right)\epsilon^{2}\right\} $.
This holds also in the general case ($|\mathbb{S}|>2$). Hence (\ref{eq:exp1})
is bounded from above by this exponential. We may therefore bound
the probability of $E_{d,\epsilon}^{(\ell)}$, for any fixed $d\in\left\{ 0,1\right\} ^{2^{k}}$,
as follows, \begin{eqnarray}
\mathbb{P}\left(E_{d,\epsilon}^{(\ell)}\right) & = & \sum_{\nu\geq\ell}\mathbb{P}\left(\left|\frac{\|\Xi^{(\nu)}\|}{\nu}-\beta\right|>\epsilon\Biggl|\nu\right)\mathbb{P}\left(\nu\right)\nonumber \\
 & \leq & 2\sum_{\nu\geq\ell}\exp\left\{ -2\nu\gamma\epsilon^{2}\right\} \mathbb{P}\left(\nu\right)\nonumber \\
 & \leq & 2\exp\left\{ -2\ell\gamma\epsilon^{2}\right\} .\label{eq:edlbound}\end{eqnarray}
Denote by $\hat{d}$ the binary vector (\ref{eq:d}) associated with
the learnt model $\mathcal{M}_{k}$ (which is based on a random training
sequence $x^{(m)}).$ We are interested in the probability of the
event $E_{\hat{d},\epsilon}^{(\ell)}$ that after learning, the selection
rule $\Gamma_{\hat{d}}$ picks a subsequence $\Xi^{(\nu)}$ from $X^{(n)}$
of length $\nu\geq\ell$ which is biased away from $\beta$ by an
amount greater than $\epsilon$. 

Denoting by $\overline{A}_{r}^{(k)}$ the complement of the set $A_{r}^{(k)}$
then we have\begin{eqnarray}
\mathbb{P}\left(E_{\hat{d},\epsilon}^{(\ell)}\right) & = & \mathbb{P}\left(E_{\hat{d},\epsilon}^{(\ell)}\biggl|\hat{d}\in A_{r}^{(k)}\right)\mathbb{P}\left(A_{r}^{(k)}\right)+\mathbb{P}\left(E_{\hat{d},\epsilon}^{(\ell)}\biggl|\hat{d}\not\in A_{r}^{(k)}\right)\mathbb{P}\left(\overline{A}_{r}^{(k)}\right)\nonumber \\
 & = & \mathbb{P}\left(\bigcup_{d\in A_{r}^{(k)}}E_{d,\epsilon}^{(\ell)}\right)\mathbb{P}\left(A_{r}^{(k)}\right)+\mathbb{P}\left(E_{\hat{d},\epsilon}^{(\ell)}\biggl|\hat{d}\not\in A_{r}^{(k)}\right)\mathbb{P}\left(\overline{A}_{r}^{(k)}\right)\nonumber \\
 & \leq & \mathbb{P}\left(\bigcup_{d\in A_{r}^{(k)}}E_{d,\epsilon}^{(\ell)}\right)+\mathbb{P}\left(\overline{A}_{r}^{(k)}\right)\nonumber \\
 & \leq & 2\left|A_{r}^{(k)}\right|\exp\left\{ -2\ell\gamma\epsilon^{2}\right\} +\exp\left\{ -2^{k-2}\left(\frac{r}{2^{k}}-\rho^{(m)}\right)^{2}\right\} \nonumber \\
 & \leq & 2\left(\frac{e2^{k}}{r}\right)^{r}\exp\left\{ -2\ell\gamma\epsilon^{2}\right\} +\exp\left\{ -2^{k-2}\left(\frac{r}{2^{k}}-\rho^{(m)}\right)^{2}\right\} \label{eq:secondf}\end{eqnarray}
which follows from (\ref{eq:dnotA}), (\ref{eq:abou}) and (\ref{eq:edlbound}).
Note that for any $0<\delta<1$ the choice of $r$ in (\ref{eq:rr})
makes the second term in (\ref{eq:secondf}) be no larger than $\frac{\delta}{2}$.
The first term is no larger than $\frac{\delta}{2}$ if the following
holds,

\begin{equation}
\epsilon\leq\sqrt{\frac{1}{2\ell\gamma}\left(r\ln\left(\frac{e2^{k}}{r}\right)+\ln2+\ln\left(\frac{4}{\delta}\right)\right)}.\label{eq:14}\end{equation}
Substituting for $r$ in (\ref{eq:14}) the value in (\ref{eq:rr})
gives the following bound on $\epsilon$,

\[
\epsilon\leq\sqrt{\frac{1}{2\ell\gamma}\left(2^{k}\left(\sqrt{\frac{1}{2^{k-2}}\ln\left(\frac{2}{\delta}\right)}+\rho^{(m)}\right)\ln\left(\frac{e}{\left(\sqrt{\frac{1}{2^{k-2}}\ln\left(\frac{2}{\delta}\right)}+\rho^{(m)}\right)}\right)+\ln2+\ln\left(\frac{4}{\delta}\right)\right)}\]
which holds with probability at least $1-\delta$. This concludes
the proof of Theorem \ref{Th1}.

\bibliographystyle{plain}
%\bibliography{/Users/Joel/MyStuff/MyBib/jerbib/jer}

\vspace{0.2cm}
\end{document}